\documentclass[a4paper, 10pt, conference]{ieeeconf}
\usepackage{FG2026}

\FGfinalcopy 

\IEEEoverridecommandlockouts
\let\labelindent\relax
\usepackage{enumitem}
\usepackage{tikz}
\usepackage{graphicx}
\usepackage{amsmath}
\usepackage{amssymb}
\usepackage{booktabs}
\usepackage{multirow}
\usepackage{url}
\usepackage{hyperref}
\usepackage{xcolor}
\usepackage{enumitem}
\usetikzlibrary{arrows.meta, positioning, calc}
\setlist{nosep,leftmargin=*}

\def\FGPaperID{****} 

\title{\LARGE \bf
Foundation Model Embeddings Meet Blended Emotions:\\
A Multimodal Fusion Approach for the BLEMORE Challenge
}

\author{\parbox{16cm}{\centering
    {\large Masoumeh Chapariniya, Aref Farhadipour, Sarah Ebling, Volker Dellwo, Teodora Vukovic}\\[4pt]
    {\normalsize Department of Computational Linguistics, University of Z\"urich, Switzerland}\\
    {\normalsize \{masoumeh.chapariniya, aref.farhadipour, ebling.cl, volker.dellwo, teodora.vukovic2\}@uzh.ch}
}}

\begin{document}

\maketitle

\begin{abstract}
We present our system for the BLEMORE Challenge at FG~2026 on blended emotion recognition with relative salience prediction. Our approach combines six encoder families through late probability fusion: an S4D-ViTMoE face encoder adapted with soft-label KL training, frozen layer-selective Wav2Vec2 audio features, finetuned body-language encoders (TimeSformer, VideoMAE), and---for the first time in emotion recognition---Gemini~Embedding~2.0, a large multimodal model whose video embeddings produce competitive presence accuracy (ACC$_\text{P}$\,=\,0.320) from only 2~seconds of input. Three key findings emerge from our experiments: selecting prosody-encoding layers (6--12) from frozen Wav2Vec2 outperforms end-to-end finetuning (Score 0.207 vs.\ 0.161), as the non-verbal nature of BLEMORE audio makes phonetic layers irrelevant; the post-processing salience threshold~$\beta$ varies from 0.05 to 0.43 across folds, revealing that personalized expression styles are the primary bottleneck; and task-adapted encoders collectively receive 62\% of ensemble weight over general-purpose baselines. Our 12-encoder system achieves Score\,=\,0.279 (ACC$_\text{P}$\,=\,0.391, ACC$_\text{S}$\,=\,0.168) on the test set, placing 6th.
\end{abstract}

\section{INTRODUCTION}

Human emotional expression is inherently complex: people frequently experience blended emotions where multiple emotions co-occur with varying prominence~\cite{berrios2015}. The BLEMORE dataset~\cite{blemore2024} captures this complexity through 3,050 video clips from 58 actors expressing 6 basic emotions and 10 blend combinations, each annotated with relative salience (50/50, 70/30, or 30/70).

The challenge requires systems to solve two coupled problems: \textit{presence recognition} (which emotions are expressed) and \textit{salience recognition} (their relative prominence). These are evaluated via $\text{Score} = 0.5 \cdot \text{ACC}_\text{P} + 0.5 \cdot \text{ACC}_\text{S}$. Critically, the evaluation pipeline converts continuous model outputs to discrete predictions through two thresholds---a presence threshold~$\alpha$ and a salience threshold~$\beta$---which are optimized on validation data, introducing actor-dependent overfitting that we analyze in detail.

Our contributions are:
\begin{enumerate}
\item \textbf{S4D-ViTMoE adaptation for blended emotions}: We develop a complete pipeline to adapt the S4D video transformer---pretrained on categorical facial expressions---for blended emotion recognition with relative salience, including face-specific preprocessing matched to the pretraining domain and soft-label KL training that preserves blend ratio signals.
\item \textbf{Layer-selective frozen audio features}: We demonstrate that extracting prosody-sensitive intermediate layers (6--12) from Wav2Vec2-large and training lightweight MLP classifiers outperforms end-to-end finetuning by 22\%, providing practical guidance for small-scale emotion datasets.
\item \textbf{Gemini Embedding 2.0 for emotion recognition}: We explore, for the first time, large multimodal foundation model (LMM) embeddings for blended emotion recognition, showing that Gemini~Embedding~2.0 achieves competitive performance (ACC$_\text{P}$\,=\,0.317) from only 2 seconds of video.
\item \textbf{Threshold sensitivity analysis}: We quantify the instability of post-processing thresholds, showing that optimal~$\beta$ varies from 0.05 to 0.43 across validation folds, and discuss implications for evaluation methodology.
\end{enumerate}


\section{RELATED WORK}
 
\textbf{Compound and blended emotion recognition.}
Du et al.~\cite{du2014compound} established compound facial expressions as combinations of basic emotions, identifying 22 categories via Action Unit analysis.
Subsequent datasets expanded this paradigm: RAF-ML~\cite{li2019blended} introduced crowdsourced multi-label annotations for blended in-the-wild expressions; MAFW~\cite{liu2022mafw} provided the first multi-modal compound emotion dataset; and C-EXPR-DB~\cite{kollias2023cexpr} offered large-scale audiovisual annotations with AU labels, adopted in the ABAW challenge series~\cite{kollias2024abaw}.
Label distribution learning~\cite{geng2016} provides a natural framework for modeling emotion mixtures as probability distributions.
However, all prior work treats compound emotions as \emph{categories} without modeling relative prominence.
The BLEMORE dataset~\cite{blemore2024} uniquely addresses this gap with explicit salience ratio annotations (50/50, 70/30, 30/70).
 
\textbf{Video-based expression recognition.}
Self-supervised pretraining has transformed dynamic facial expression recognition (DFER). MAE-DFER~\cite{sun2023maedfer} introduced masked autoencoder pretraining on facial videos, surpassing supervised methods.
The S4D framework~\cite{s4d2024}, which we adopt as our primary visual encoder, uses dual-modal pretraining on static images and videos with Mixture-of-Adapter-Experts modules.
TimeSformer~\cite{bertasius2021timesformer} and VideoMAE~\cite{tong2022videomae} provide strong pretrained backbones that we finetune for body-level gesture recognition.
 
\textbf{Self-supervised audio features for emotion.}
Self-supervised speech models are now dominant feature extractors for speech emotion recognition~\cite{wagner2023dawn}.
Layer-wise analyses~\cite{pasad2023comparative} reveal that middle transformer layers (6--12) encode prosody and paralinguistics while upper layers specialize for phonetics---a finding we exploit through selective layer extraction.
The SUPERB benchmark~\cite{yang2021superb} established the paradigm of frozen SSL features with lightweight classifiers.
For multimodal fusion, HiCMAE~\cite{sun2024hicmae} combines hierarchical contrastive learning with masked autoencoders for audio-visual emotion recognition.
 
\textbf{Foundation model embeddings for emotion.}
Recent work explores large multimodal models (LMMs) for emotion understanding. GPT-4V has been benchmarked across 21 emotion datasets~\cite{lian2024gpt4v}, while Emotion-LLaMA~\cite{cheng2024emotionllama} integrates multimodal encoders into a unified LLM.
ImageBind~\cite{girdhar2023imagebind} provides joint embeddings across modalities.
However, using LMM \emph{embeddings} as features for fine-grained blended emotion classification remains unexplored.
We address this gap by incorporating Gemini Embedding 2.0~\cite{gemini_embedding2025} into our ensemble.
\section{METHOD}

\begin{figure}[t]
\centering
\resizebox{\columnwidth}{!}{%
\begin{tikzpicture}[
    node distance=0.35cm,
    every node/.style={font=\footnotesize},
    box/.style={draw, rounded corners=2pt, minimum height=0.5cm, align=center, line width=0.4pt},
    enc/.style={box, minimum width=1.28cm, minimum height=0.75cm},
    wide/.style={box, minimum width=7cm},
    prep/.style={box, fill=gray!6, minimum width=1.28cm, minimum height=0.55cm},
    arr/.style={-{Stealth[length=2pt]}, gray!60, line width=0.3pt},
]

\node[enc, fill=purple!12] (s4d) {\scriptsize\textbf{S4D-ViTMoE}\\\tiny Face};
\node[enc, fill=teal!12, right=0.12cm of s4d] (ts) {\scriptsize\textbf{TimeSformer}\\\tiny Body};
\node[enc, fill=teal!12, right=0.12cm of ts] (vmae) {\scriptsize\textbf{VideoMAE}\\\tiny Body};
\node[enc, fill=orange!12, right=0.12cm of vmae] (w2v) {\scriptsize\textbf{Wav2Vec2}\\\tiny Layers 6--12};
\node[enc, fill=blue!12, right=0.12cm of w2v] (gem) {\scriptsize\textbf{Gemini 2.0}\\\tiny LMM embed.};

\node[prep, above=0.35cm of s4d] (pr1) {\tiny MTCNN\\\tiny Face Crops};
\node[prep, above=0.35cm of ts] (pr2) {\tiny Body\\\tiny Frames};
\node[prep, above=0.35cm of vmae] (pr3) {\tiny Body\\\tiny Frames};
\node[prep, above=0.35cm of w2v] (pr4) {\tiny Audio\\\tiny wav};
\node[prep, above=0.35cm of gem] (pr5) {\tiny First 2s\\\tiny Video};

\coordinate (midtop) at ($(pr1.north)!0.5!(pr5.north)$);
\node[wide, fill=gray!12, above=0.35cm of midtop] (input) {\textbf{Video clip}};

\draw[arr] (input.south -| pr1) -- (pr1.north);
\draw[arr] (input.south -| pr2) -- (pr2.north);
\draw[arr] (input.south -| pr3) -- (pr3.north);
\draw[arr] (input.south -| pr4) -- (pr4.north);
\draw[arr] (input.south -| pr5) -- (pr5.north);

\draw[arr] (pr1.south) -- (s4d.north);
\draw[arr] (pr2.south) -- (ts.north);
\draw[arr] (pr3.south) -- (vmae.north);
\draw[arr] (pr4.south) -- (w2v.north);
\draw[arr] (pr5.south) -- (gem.north);

\node[below=0.02cm of s4d, font=\tiny, gray] (cl1) {KL head};
\node[below=0.02cm of ts, font=\tiny, gray] (cl2) {KL head};
\node[below=0.02cm of vmae, font=\tiny, gray] (cl3) {KL head};
\node[below=0.02cm of w2v, font=\tiny, gray] (cl4) {MLP};
\node[below=0.02cm of gem, font=\tiny, gray] (cl5) {MLP};

\draw[arr] (s4d.south) -- (cl1.north);
\draw[arr] (ts.south) -- (cl2.north);
\draw[arr] (vmae.south) -- (cl3.north);
\draw[arr] (w2v.south) -- (cl4.north);
\draw[arr] (gem.south) -- (cl5.north);

\node[box, fill=gray!6, minimum width=0.9cm, below=0.25cm of cl1, font=\tiny] (p1) {$\mathbf{p}_\text{face}$};
\node[box, fill=gray!6, minimum width=0.9cm, below=0.25cm of cl2, font=\tiny] (p2) {$\mathbf{p}_\text{ts}$};
\node[box, fill=gray!6, minimum width=0.9cm, below=0.25cm of cl3, font=\tiny] (p3) {$\mathbf{p}_\text{vm}$};
\node[box, fill=gray!6, minimum width=0.9cm, below=0.25cm of cl4, font=\tiny] (p4) {$\mathbf{p}_\text{aud}$};
\node[box, fill=gray!6, minimum width=0.9cm, below=0.25cm of cl5, font=\tiny] (p5) {$\mathbf{p}_\text{gem}$};

\draw[arr] (cl1.south) -- (p1.north);
\draw[arr] (cl2.south) -- (p2.north);
\draw[arr] (cl3.south) -- (p3.north);
\draw[arr] (cl4.south) -- (p4.north);
\draw[arr] (cl5.south) -- (p5.north);

\coordinate (pmid) at ($(p1.south)!0.5!(p5.south)$);
\node[font=\tiny, gray, below=0.15cm of pmid] (blabel) {+ HiCMAE, WavLM, ImageBind, CLIP, DINOv2, VideoSwin};

\node[wide, fill=yellow!15, minimum height=0.6cm, below=0.35cm of blabel] (fusion) {\textbf{Weighted late fusion} \quad {\scriptsize $\mathbf{p}_\text{fused} = \textstyle\sum_m w_m \mathbf{p}_m$}};

\draw[arr] (p1.south) -- ++(0,-0.35) -| ([xshift=-2cm]fusion.north);
\draw[arr] (p2.south) -- ++(0,-0.35) -| ([xshift=-1cm]fusion.north);
\draw[arr] (p3.south) -- ++(0,-0.35) -| (fusion.north);
\draw[arr] (p4.south) -- ++(0,-0.35) -| ([xshift=1cm]fusion.north);
\draw[arr] (p5.south) -- ++(0,-0.35) -| ([xshift=2cm]fusion.north);

\node[wide, fill=pink!12, minimum height=0.6cm, below=0.25cm of fusion] (post) {\textbf{Post-processing} \quad {\scriptsize Top-2 $\to$ $\alpha$ (presence) $\to$ $\beta$ (salience)}};
\draw[arr] (fusion) -- (post);

\node[wide, fill=green!12, below=0.25cm of post] (out) {\textbf{Prediction} \quad {\scriptsize e.g., \{anger: 70, fear: 30\}}};
\draw[arr] (post) -- (out);

\end{tikzpicture}
}
\caption{Overview of the multimodal ensemble pipeline. A video clip is processed by five encoder families operating on different input representations: face-cropped frames (S4D-ViTMoE), body frames (TimeSformer, VideoMAE), audio waveforms (Wav2Vec2 layers 6--12), short video clips (Gemini Embedding 2.0), and pre-extracted features from challenge baselines. Each encoder produces a 6-dimensional probability vector over emotions, which are combined through optimized weighted averaging. The fused probabilities are converted to discrete predictions via threshold-based post-processing ($\alpha$ for presence, $\beta$ for salience).}
\label{fig:pipeline}
\end{figure}
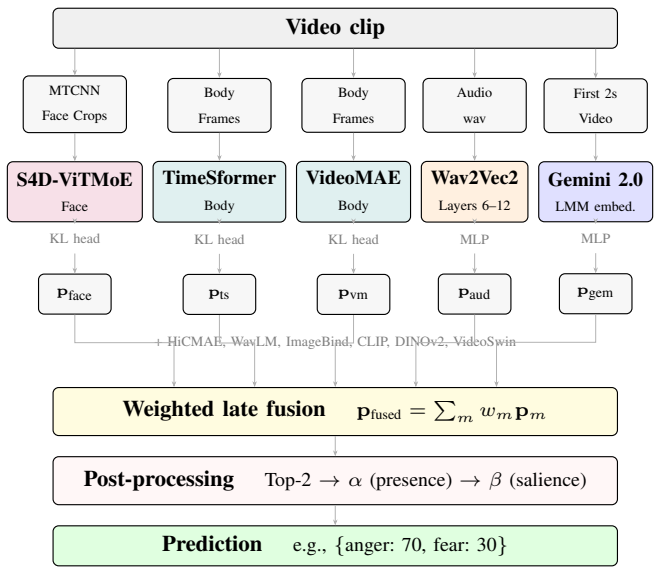

Our system consists of six encoder families whose probability outputs are combined through late fusion (Fig.~\ref{fig:pipeline}).

\subsection{Visual Encoder: S4D-ViTMoE (Face)}

We adapt the S4D framework~\cite{s4d2024} for blended emotion recognition. S4D uses a Vision Transformer (ViT-Base) with Mixture-of-Experts (MoE) adapters, pretrained on VoxCeleb2 (talking-head videos) and AffectNet (static facial expressions) for categorical expression recognition. Our adaptation involves three key modifications.

\textbf{Face-aligned preprocessing.} We extract face crops using MTCNN at 30fps, aligned and resized to $160{\times}160$, matching the VoxCeleb2 pretraining domain. Since S4D was pretrained on talking-head videos and static face images, face-cropped inputs preserve the feature distributions learned during pretraining, while full-frame inputs would introduce background and body regions outside the pretrained distribution. We capture body-level cues through separate dedicated encoders (TimeSformer, VideoMAE) rather than compromising the face encoder's input domain.

\textbf{Soft-label KL training.} The original S4D uses cross-entropy for categorical emotions. We replace this with KL divergence on soft labels encoding the exact blend ratio: $\mathbf{y}_\text{soft} = [0.7, 0, 0.3, 0, 0, 0]$ for a 70/30 anger-fear blend. This preserves the salience signal that cross-entropy would discard.

\textbf{Selective finetuning.} Backbone blocks 0--5 are frozen (${\sim}$38M params); only MoE adapter layers in blocks 6--11 and the classification head are trained (${\sim}$12M params). The model processes 16 frames with tubelet embedding (tubelet size\,=\,2), producing 800 spatiotemporal tokens (8 temporal $\times$ 100 spatial).

\textbf{Temporal sampling strategy.} With a sampling rate of 4, the model selects 16 frames spanning a 64-frame window (${\sim}$2.1s at 30fps). During training, the starting position is selected \emph{randomly} within the video, so each epoch sees a different temporal crop of the same clip---serving as an implicit data augmentation that exposes the model to varying phases of the emotional expression. During validation and testing, we use a \emph{deterministic} multi-clip strategy: the video is divided into 2 temporal segments with evenly spaced start positions, and 2 spatial crops are taken per segment, yielding 4 clips per video. The softmax probabilities across all 4 clips are averaged to produce the final prediction, reducing sensitivity to the exact temporal window.

\subsection{Audio Encoder: Layer-Selective Wav2Vec2}

A distinctive characteristic of BLEMORE audio is that actors produce \emph{non-linguistic vocalizations}---laughs, cries, gasps, and growls---rather than spoken words~\cite{blemore2024}. Since Wav2Vec2 \cite{wav2vec} was pretrained for ASR, its upper transformer layers are optimized for phonetic encoding, which is irrelevant for non-verbal audio. Layer-wise analyses~\cite{pasad2023comparative} show that \emph{middle layers (6--12) encode prosody, rhythm, and paralinguistic properties}---precisely the features that carry emotional information in non-verbal vocalizations: pitch contour, vocal effort, and temporal rhythm.

\textbf{Frozen layer-selective features.} We extract hidden states from layers 6--12 of (frozen), average them, and aggregate per-frame features with temporal statistics (mean, std, percentiles) over 3 segments, yielding 7,168-dim vectors classified by an MLP (1024$\rightarrow$512$\rightarrow$6, BatchNorm, dropout\,=\,0.3).

\textbf{End-to-end finetuning.} For comparison, we finetune with attentive pooling and discriminative learning rates ($10^{-5}$ backbone, $10^{-3}$ head).

The frozen layer-selective approach (Score\,=\,0.207) dramatically outperforms finetuning (Score\,=\,0.161)---a 22\% gap. Finetuning 95M parameters on only 2,456 samples leads to severe overfitting, while the frozen approach preserves rich pretrained representations with only a lightweight classifier to learn. This underscores that for small non-verbal emotion datasets, \emph{selecting the right pretrained layers matters more than end-to-end adaptation}.

\subsection{Body Language Encoders}

Research has shown that body movements can improve emotion recognition accuracy when combined with facial expressions~\cite{luo2020bodylanguage}, as posture, gestures, and limb dynamics convey emotional intensity that faces alone may not fully capture---particularly for blended emotions where the dominant emotion may manifest more strongly in body language than in subtle facial muscle changes. To capture these gestural and postural cues, we finetune two video transformers on full-body frames (no face cropping), providing representations complementary to our face-specific S4D encoder.

\textbf{VideoMAE}~\cite{tong2022videomae}: A masked video autoencoder that reconstructs 90--95\% masked tube tokens, learning motion-sensitive representations from Kinetics-400. We finetune on 16 body-cropped frames with soft-label KL training. 

\textbf{TimeSformer}~\cite{bertasius2021timesformer}: A divided space-time attention transformer that factorizes spatial and temporal attention within each block, enabling efficient modeling of long-range temporal dependencies. We finetune from a Kinetics-600 checkpoint on 8 frames.

Both use the same soft-label KL training, discriminative learning rates ($10^{-5}$ backbone, $10^{-3}$ head), and early stopping strategy. While individually weaker than S4D, these body-level encoders contribute meaningfully to fusion by capturing emotion cues inaccessible from face crops alone.

\subsection{LMM Embeddings: Gemini Embedding 2.0}

We explore a novel approach: using embeddings from a large multimodal model (LMM) as features for emotion recognition. Specifically, we use Gemini~Embedding~2.0~\cite{gemini_embedding2025}, which produces 3,072-dim embeddings from multimodal inputs including video.

Due to the model's current 2-second video input limit, we extract embeddings from only the first 2 seconds of each clip. Despite this severe temporal constraint (BLEMORE videos are 3--8 seconds), the Gemini embeddings achieve ACC$_\text{P}$\,=\,0.317 and Score\,=\,0.231---competitive with domain-specific encoders trained on the full video. This suggests that LMM embeddings capture rich emotional semantics even from brief temporal windows, and represents a promising direction for emotion recognition research.

\subsection{Pre-Extracted Baseline Features}

We additionally train MLP classifiers on pre-extracted features provided by the challenge organizers: HiCMAE~\cite{sun2024hicmae} (cross-modal audio-visual), WavLM~\cite{chen2022wavlm} (speech), ImageBind~\cite{girdhar2023imagebind}, CLIP, DINOv2, DINOv3, and VideoSwin~\cite{liu2022videoswin}. These provide complementary representations that improve ensemble diversity.

\subsection{Late Fusion Ensemble}

Given $M$ encoders with softmax probability vectors $\mathbf{p}_m \in \mathbb{R}^6$, the fused prediction is $\mathbf{p}_\text{fused} = \sum_{m=1}^{M} w_m \cdot \mathbf{p}_m$, with $\sum w_m = 1$. Weights are optimized via constrained grid search on validation folds, maximizing the combined score. Table~\ref{tab:weights} shows the learned weights for our two ensemble configurations.

\subsection{Post-Processing}

Following the official BLEMORE pipeline~\cite{blemore2024}, fused probabilities are converted to discrete predictions via: (1)~top-2 masking, (2)~suppressing probabilities below a presence threshold~$\alpha$, (3)~collapsing to single emotion if neutral appears, and (4)~assigning 50/50 salience if $|p_1 - p_2| \leq \beta$, else 70/30. Thresholds $(\alpha, \beta)$ are selected via grid search on validation data (Section~\ref{sec:threshold}).

\section{EXPERIMENTS}
\subsection{Setup}
The BLEMORE training set contains 2,456 clips from 43~actors (5-fold actor-disjoint CV); the test set has 594 clips from 15 unseen actors. The training distribution is 46\% single emotions, 18\% 50/50 blends, and 36\% 70/30 blends. S4D is trained for 100 epochs (batch~16, lr~$10^{-4}$, cosine schedule). Audio MLPs use 500 epochs with patience-80 early stopping. Body encoders use 30 epochs with patience-10. Gemini embeddings are extracted via API; MLPs are trained with the same protocol as audio.
\subsection{Individual Encoder Results}
Table~\ref{tab:individual} presents all encoder results. Our S4D-ViTMoE achieves the best single-encoder presence accuracy (0.340), outperforming all baselines including ImageBind (0.290) and VideoMAEv2 (0.273). Gemini~Embedding~2.0 achieves ACC$_\text{P}$\,=\,0.317 from only 2~seconds of video---surpassing domain-specific encoders that process full clips, demonstrating that LMM embeddings capture rich emotional semantics.

\subsection{Fusion and Test Results}

Table~\ref{tab:fusion} compares our submissions against the BLEMORE baselines on the test set. Our 12-encoder ensemble with Gemini achieves ACC$_\text{P}$\,=\,0.391, substantially outperforming the best baseline presence (VideoMAEv2\,+\,HuBERT, 0.332). Adding Gemini embeddings improved our test score from 0.262 to 0.279 (+6.5\%), with presence accuracy increasing from 0.357 to 0.391---confirming that LMM embeddings provide complementary information that generalizes to unseen actors. Notably, Gemini receives the highest weight in the 12-encoder fusion (0.192, Table~\ref{tab:weights}), suggesting it captures emotional semantics unavailable to domain-specific encoders.

\begin{table}[t]
\caption{Individual encoder results (5-fold CV). M\,=\,modality. \textsuperscript{$\dagger$}Baselines from~\cite{blemore2024}. \textsuperscript{$\ddagger$}Our finetuned models.}
\label{tab:individual}
\centering
\small
\setlength{\tabcolsep}{3pt}
\begin{tabular}{@{}llccc@{}}
\toprule
\textbf{Encoder} & \textbf{M} & \textbf{ACC$_\text{P}$} & \textbf{ACC$_\text{S}$} & \textbf{Score} \\
\midrule
\multicolumn{5}{l}{\textit{Our adapted/finetuned models}\textsuperscript{$\ddagger$}} \\
S4D-ViTMoE (face) & V & $.340 {\scriptstyle\pm .024}$ & $.140 {\scriptstyle\pm .018}$ & $.240$ \\
Gemini Embed.\ 2.0 (2s) & AV & $.320 {\scriptstyle\pm .033}$ & $.137 {\scriptstyle\pm .009}$ & $.223$ \\
VideoMAE body (ft) & V & $.291 {\scriptstyle\pm .016}$ & $.144 {\scriptstyle\pm .026}$ & $.218$ \\
Wav2Vec2 frozen+MLP-1024 & A & $.294 {\scriptstyle\pm .025}$ & $.120 {\scriptstyle\pm .018}$ & $.207$ \\
TimeSformer body (ft) & V & $.259 {\scriptstyle\pm .031}$ & $.131 {\scriptstyle\pm .015}$ & $.195$ \\
Wav2Vec2 frozen+MLP-512 & A & $.264 {\scriptstyle\pm .031}$ & $.104 {\scriptstyle\pm .016}$ & $.184$ \\
Wav2Vec2 finetuned E2E & A & $.234 {\scriptstyle\pm .026}$ & $.088 {\scriptstyle\pm .020}$ & $.161$ \\
\midrule
\multicolumn{5}{l}{\textit{BLEMORE baselines (pre-extracted features)}\textsuperscript{$\dagger$}} \\
HiCMAE & AV & $.298 {\scriptstyle\pm .025}$ & $.180 {\scriptstyle\pm .036}$ & $.239$ \\
ImageBind & V & $.290 {\scriptstyle\pm .028}$ & $.130 {\scriptstyle\pm .008}$ & $.210$ \\
WavLM & A & $.265 {\scriptstyle\pm .027}$ & $.121 {\scriptstyle\pm .012}$ & $.193$ \\
VideoMAEv2\textsuperscript{$\dagger$} & V & $.273 {\scriptstyle\pm .025}$ & $.106 {\scriptstyle\pm .014}$ & $.190$ \\
\bottomrule
\end{tabular}
\vspace{-3mm}
\end{table}
\begin{table}[t]
\caption{Fusion results compared to BLEMORE baselines~\cite{blemore2024}. Val\,=\,5-fold CV. Test\,=\,competition server.}
\label{tab:fusion}
\centering
\small
\setlength{\tabcolsep}{3pt}
\begin{tabular}{@{}lccc@{}}
\toprule
\textbf{Configuration} & \textbf{ACC$_\text{P}$} & \textbf{ACC$_\text{S}$} & \textbf{Score} \\
\midrule
\multicolumn{4}{l}{\textit{BLEMORE baselines (test)}~\textsuperscript{$\dagger$}} \\
VideoMAEv2 + HuBERT\textsuperscript{$\dagger$} & .332 & .114 & .223 \\
ImageBind + WavLM\textsuperscript{$\dagger$} & .327 & .114 & .221 \\
HiCMAE\textsuperscript{$\dagger$} & .268 & \textbf{.180} & .224 \\
\midrule
\multicolumn{4}{l}{\textit{Our submissions (test)}} \\
S4D + Wav2Vec2 & .327 & .159 & .243 \\
9-encoder ensemble & .357 & .168 & .262 \\
\textbf{12-enc + Gemini} & $\mathbf{.391}$ & .168 & $\mathbf{.279}$ \\
\midrule
\multicolumn{4}{l}{\textit{Our validation (5-fold CV)}} \\
S4D face only & .340 & .140 & .240 \\
S4D + Wav2Vec2 & .357 & .175 & .266 \\
9-encoder & .414 & .205 & .309 \\
12-encoder + Gemini & .418 & .204 & .311 \\
\bottomrule
\end{tabular}
\vspace{-3mm}
\end{table}
\subsection{Threshold Sensitivity Analysis}
\label{sec:threshold}
\begin{table}[t]
\caption{Optimized encoder weights for two ensemble configurations. Weights are averaged across 5-fold cross-validation. Dashes indicate encoders not included in that configuration.}
\label{tab:weights}
\centering
\small
\setlength{\tabcolsep}{4pt}
\begin{tabular}{@{}llcc@{}}
\toprule
& \textbf{Encoder} & \textbf{9-enc} & \textbf{12-enc} \\
\midrule
\multirow{3}{*}{\rotatebox{90}{\scriptsize Ours}}
& S4D-ViTMoE (face)  & .094 & .117 \\
& Wav2Vec2 (audio)    & .170 & .111 \\
& Gemini Embed.\ 2.0  & ---  & \textbf{.192} \\
& TimeSformer (body)  & ---  & .090 \\
& VideoMAE (body)     & ---  & .110 \\
\midrule
\multirow{4}{*}{\rotatebox{90}{\scriptsize Baseline}}
& HiCMAE              & \textbf{.261} & .103 \\
& WavLM               & .156 & .124 \\
& ImageBind           & .050 & .079 \\
& CLIP                & .071 & --- \\
& DINOv2              & .041 & --- \\
& DINOv3              & .092 & .042 \\
& VideoSwin           & .064 & .032 \\
\midrule
\multicolumn{2}{l}{Val ACC$_\text{P}$ / ACC$_\text{S}$ / Score}
& .386 / .150 / .268
& .393 / .181 / .287 \\
\multicolumn{2}{l}{Test ACC$_\text{P}$ / ACC$_\text{S}$ / Score}
& .357 / .168 / .262
& \textbf{.391} / .168 / \textbf{.279} \\
\bottomrule
\end{tabular}
\vspace{-3mm}
\end{table}
Following the official BLEMORE protocol~\cite{blemore2024}, continuous outputs are discretized via presence threshold~$\alpha$ and salience threshold~$\beta$, selected by grid search on validation data. Across our ensemble configurations, we explored per-fold optimization with averaging, decoupled sequential search ($\alpha$ for presence first, then $\beta$ for salience), and best-fold selection for test.

A critical finding is the instability of~$\beta$: per-fold optimal values range from 0.05 to 0.43---an 8$\times$ variation representing opposite strategies---indicating that $\beta$ is fundamentally actor-dependent. In contrast, $\alpha$ remains stable (0.10--0.14). This explains the persistent val$\rightarrow$test gap (val 0.311 $\rightarrow$ test 0.279) and represents a fundamental challenge for threshold-based evaluation, as acknowledged in~\cite{blemore2024}.

\section{DISCUSSION AND FUTURE WORK}

\textbf{LMM embeddings and encoder weights.} Gemini~Embedding~2.0 achieves ACC$_\text{P}$\,=\,0.317 from only 2~seconds of video and receives the highest fusion weight (0.192, Table~\ref{tab:weights}). Its limited salience accuracy (0.145) likely reflects the 2-second input constraint, as blended clips tend to be longer than single-emotion clips~\cite{blemore2024} and salience cues may emerge later. Overall, our adapted models collectively receive 62\% of ensemble weight versus 38\% for pre-extracted baselines, demonstrating that task-specific adaptation outperforms general-purpose features for this specialized task.

\textbf{Frozen features and personalized expression.} The gap between frozen layer-selective Wav2Vec2 and end-to-end finetuning highlights that for small non-verbal datasets, selecting task-relevant pretrained layers matters more than architectural adaptation. More broadly, emotional expression is inherently personal---individuals convey the same blend with idiosyncratic dynamics---which is the root cause of the $8\times$ $\beta$ variation across folds and the persistent val$\rightarrow$test gap.

\textbf{Future directions.} Promising paths include: (i)~full-length LMM video inputs with prompt engineering for emotion; (ii)~specialized models for blended emotions, which are systematically longer and may require different temporal modeling; (iii)~continuous salience regression to eliminate threshold sensitivity; (iv)~actor-adaptive normalization and threshold-free prediction; and (v)~cross-dataset transfer from larger expression datasets (MAFW, Aff-Wild2).

\section{CONCLUSION}

We presented a multimodal ensemble for blended emotion recognition combining S4D-ViTMoE, layer-selective Wav2Vec2, body encoders, and---for the first time in emotion recognition---LMM embeddings (Gemini~Embedding~2.0). Our system achieves Score\,=\,0.279 on the BLEMORE test set (ACC$_\text{P}$\,=\,0.391), substantially outperforming the best baseline (0.332). Three key insights emerge: frozen layer-selective features outperform finetuning on small non-verbal data, demonstrating that choosing the right pretrained layers matters more than end-to-end adaptation; LMM embeddings achieve competitive emotion recognition from just 2~seconds of video, opening a new direction for affective computing; and the salience threshold exhibits fundamental actor-dependent instability, revealing that personalized expression styles---not model capacity---are the primary bottleneck for blended emotion recognition.

\textbf{Code}: \url{https:github.com/Mchapariniya/blemore-multimodal}

{\small
\bibliographystyle{ieee}

\begin{thebibliography}{22}

\bibitem{berrios2015}
R.~Berrios, P.~Totterdell, and S.~Kellett, ``Eliciting mixed emotions: A meta-analysis comparing models, types, and measures,'' \textit{Frontiers in Psychology}, vol.~6, 2015.

\bibitem{blemore2024}
T.~Lachmann, A.~Israelsson, C.~Tornberg, T.~Saghinadze, M.~Balazia, P.~M\"uller, and P.~Laukka, ``Not all blends are equal: The BLEMORE dataset of blended emotion expressions with relative salience annotations,'' \textit{arXiv:2601.13225}, 2025.

\bibitem{du2014compound}
S.~Du, Y.~Tao, and A.~M.~Martinez, ``Compound facial expressions of emotion,'' \textit{PNAS}, vol.~111, no.~15, pp.~E1454--E1462, 2014.

\bibitem{li2019blended}
S.~Li and W.~Deng, ``Blended emotion in-the-wild: Multi-label facial expression recognition using crowdsourced annotations and deep locality feature learning,'' \textit{IJCV}, vol.~127, no.~6--7, pp.~884--906, 2019.

\bibitem{liu2022mafw}
Y.~Liu \textit{et al.}, ``MAFW: A large-scale, multi-modal, compound affective database for dynamic facial expression recognition in the wild,'' in \textit{Proc.\ ACM MM}, 2022.

\bibitem{kollias2023cexpr}
D.~Kollias, ``Multi-label compound expression recognition: C-EXPR database \& network,'' in \textit{Proc.\ CVPR}, 2023, pp.~5589--5598.

\bibitem{kollias2024abaw}
D.~Kollias \textit{et al.}, ``The 7th ABAW competition: Multi-task learning and compound expression recognition,'' in \textit{Proc.\ ECCV Workshops}, 2024.

\bibitem{geng2016}
X.~Geng, ``Label distribution learning,'' \textit{IEEE TKDE}, vol.~28, no.~7, pp.~1734--1748, 2016.

\bibitem{sun2023maedfer}
L.~Sun \textit{et al.}, ``MAE-DFER: Efficient masked autoencoder for self-supervised dynamic facial expression recognition,'' in \textit{Proc.\ ACM MM}, 2023, pp.~6110--6121.

\bibitem{s4d2024}
Y.~Chen \textit{et al.}, ``Static for dynamic: Towards a deeper understanding of dynamic facial expressions using static expression data,'' \textit{IEEE Trans.\ Affect.\ Comput.}, 2025.

\bibitem{bertasius2021timesformer}
G.~Bertasius, H.~Wang, and L.~Torresani, ``Is space-time attention all you need for video understanding?'' in \textit{Proc.\ ICML}, 2021.

\bibitem{tong2022videomae}
Z.~Tong \textit{et al.}, ``VideoMAE: Masked autoencoders are data-efficient learners for self-supervised video pre-training,'' in \textit{Proc.\ NeurIPS}, 2022.

\bibitem{wagner2023dawn}
J.~Wagner \textit{et al.}, ``Dawn of the transformer era in speech emotion recognition: Closing the valence gap,'' \textit{IEEE TPAMI}, vol.~45, no.~9, pp.~10745--10759, 2023.
\bibitem{wav2vec}
A.~Baevski, Y.~Zhou, A.~Mohamed, and M.~Auli, ``wav2vec 2.0: A framework for self-supervised learning of speech representations,'' in \textit{Proc.\ NeurIPS}, 2020.
\bibitem{pasad2023comparative}
A.~Pasad, B.~Shi, and K.~Livescu, ``Comparative layer-wise analysis of self-supervised speech models,'' in \textit{Proc.\ ICASSP}, 2023.

\bibitem{yang2021superb}
S.~Yang \textit{et al.}, ``SUPERB: Speech processing universal performance benchmark,'' in \textit{Proc.\ Interspeech}, 2021, pp.~1194--1198.

\bibitem{sun2024hicmae}
L.~Sun \textit{et al.}, ``HiCMAE: Hierarchical contrastive masked autoencoder for self-supervised audio-visual emotion recognition,'' \textit{Information Fusion}, vol.~108, art.~102382, 2024.

\bibitem{lian2024gpt4v}
Z.~Lian \textit{et al.}, ``GPT-4V with emotion: A zero-shot benchmark for generalized emotion recognition,'' \textit{Information Fusion}, vol.~108, art.~102367, 2024.

\bibitem{cheng2024emotionllama}
Z.~Cheng \textit{et al.}, ``Emotion-LLaMA: Multimodal emotion recognition and reasoning with instruction tuning,'' in \textit{Proc.\ NeurIPS}, 2024.

\bibitem{girdhar2023imagebind}
R.~Girdhar \textit{et al.}, ``ImageBind: One embedding space to bind them all,'' in \textit{Proc.\ CVPR}, 2023.

\bibitem{gemini_embedding2025}
J.~Lee \textit{et al.}, ``Gemini embedding: Generalizable embeddings from Gemini,'' \textit{arXiv:2503.07891}, 2025.

\bibitem{luo2020bodylanguage}
Y.~Luo \textit{et al.}, ``Learning facial expression and body gesture visual information for video emotion recognition,'' \textit{Expert Systems with Applications}, vol.~237, art.~121419, 2024.



\bibitem{chen2022wavlm}
S.~Chen \textit{et al.}, ``WavLM: Large-scale self-supervised pre-training for full stack speech processing,'' \textit{IEEE J-STSP}, vol.~16, no.~6, pp.~1505--1518, 2022.

\bibitem{liu2022videoswin}
Z.~Liu \textit{et al.}, ``Video Swin Transformer,'' in \textit{Proc.\ CVPR}, 2022, pp.~3192--3201.

\bibitem{israelsson2023}
A.~Israelsson, A.~Seiger, and P.~Laukka, ``Blended emotions can be accurately recognized from dynamic facial and vocal expressions,'' \textit{J.\ Nonverbal Behavior}, vol.~47, pp.~1--18, 2023.

\end{thebibliography}

}
\end{document}